\documentclass{article}

\usepackage[final]{corl_2020} 
\usepackage{graphicx}
\usepackage{wrapfig}
\graphicspath{{./figs/}}

\title{Autonomous soft hand grasping - Literature review}

\usepackage{multirow}
\usepackage{booktabs}

 \usepackage{amsmath}
  \usepackage{amssymb}
  \usepackage{amsfonts}

\usepackage{algorithm}
\usepackage{algpseudocode}
\algrenewcommand{\algorithmicrequire}{\textbf{Input:~~}}
\algrenewcommand{\algorithmicensure}{\textbf{Output:}}
\algrenewcommand{\algorithmiccomment}[1]{\qquad\hfill~\hspace*{-5ex}\textit{// #1}}

\author{
  Tai Hoang\\
  Technical University of Munich\\
  Munich, Germany\\
  \texttt{t.hoang@tum.de} \\
}

\begin{document}
\maketitle


\begin{abstract}
Autonomous grasping remains challenging as unlike humans, robots do not possess a sophisticated sensing nor delicate interaction capability with the real environment. Among other efforts that tried to close the gap between them, anthropomorphic robotic hands is the most prominent direction. However, exactly following human hand design might be unnecessary as it will significantly increase the mechanical complexity and hence make it less economically feasible. Recently, soft robotic hands, a new trend has emerged, aiming to make the design adequately complex and affordable while requiring much less effort to control. In the first part of this article, we will lay out several prominent designs in this direction and their applications in real world scenarios. Having a suitable hardware simplified the complexity of software designing. However, manually controlling the hand for one task requires a significantly large amount of time and effort and doing it repeatedly is unsurprisingly tedious. Therefore, in the second part, we will show some recent techniques for soft hand autonomous control. We start by briefly discussing the analytic methods that mainly exploit the hand dynamic information. Then, data-driven approaches will be our main focus. It is the trending research topic for soft hand grasping in recent years as it has shown a high performance when dealing with a large number of various objects.

\end{abstract}



\section{Introduction}
Human hands show a great capability of grasping in the sense of reliability and dexterity. Designing a mechanical hand that can do almost anything like a human hand has always been the engineer's greatest ambition \cite{hand_challenges}. The road to achieve this goal is long and difficult, even up till this point, with more than a century of development and hundreds of designs has come out \cite{century_of_hands}, there still exists some limitations and open problems. 

In autonomous robot design, there is a branch of research where people often neglect the importance of the end-effector, try to put the focus more on the upper part of the robot such as their arm or body. The mechanical design of the hand is thus significantly simpler, a rigid gripper attached to the end of the robot's arm. Apparently, a simple gripper only allows the robot to grasp limited objects with specific shape and size. While in the other direction, the aim is designing a robot hand as much flexible as a human hand as possible. In the early 2000s, a lot of designs followed the human hand structure with high mechanical complexity, (i.e fully actuated with many degrees of freedom) \cite{century_of_hands}. However, such a rigid hand is costly and more prone to damage. In order to reduce the risks, some compliance was integrated into the mechanical parts differently \cite{rbo_thesis}. Though compliant hands were a notable turning point, fully actuated hands might be unnecessarily complex. In fact, in the DARPA Robotics Challenge \cite{DARPA}, a renowned international robotic competition, none of the robots followed the fully actuated scheme, whereas 15/25 hands were underactuated with only three or four fingers. This showed that for practical applications, minimalistic design is a more appropriate principle \cite{hand_challenges}. 

SDM hand \cite{SDM} was one of the first prototypes following this design principle. The innovation is the elastomer fingers it uses, still possess a high number of degrees of freedom though, can be passively controlled by using only one single actuator through the tendon-pulley system. Similarly, Pisa/IIT SoftHand \cite{PSH} is another notable underactuated hand. Its design complies with the anthropomorphic form and the postural synergies hypothesis. The hypothesis states that most of the grasping postures can be explained by a small set of basis called eigengrasp space \cite{eigengrasp}. In fact, 80\% of postures can be described by the first two synergies, and 87\% with only the first three \cite{eigengrasp_fact}. Inspired by this, they design a tendon routing in a way that enables the robot to recreate these hand postures under one \cite{PSH} or just two actuators \cite{PSH2}. Apart from the tendon-driven system, RBO \cite{rbo_thesis} and its recent version RBO 2 is an anthropomorphic and entirely soft continuum hand. The whole hand is made of silicone material and actuated via the pneumatic mechanism attached on five fingers and two in the palm for opposing the thumb purpose. 

Designing a control strategy for robust grasping remains challenging. Though lots of effort has been put throughout a few decades, the problem can only be solved partially due to the limitation in hardware and (or) computational power. The approaches trying to address this can be divided into two main categories, analytic and empirical methods \cite{data_driven_survey}. Analytic methodology aims to exploit the dynamic behaviours of the robot hand and the targeted object, formalize them into a constrained optimization problem \cite{analytic_grasping} and find a solution (either in exact or approximate form) for it. To this end, lots of research uses the concept of “independent contact regions” (ICRs) \cite{icr} as a way to measure grasp quality. ICRs is a set of regions in which as long as the contact points stay in these regions, the grasp will remain stable. Due to the lack of compliant components, most of the early analytical approaches were applied on rigid robotic hands and in this specific setting, a stable grasp usually means a collision-free path. This along with the rigidity and fragility of the hand greatly increase the complexity of constraints, forms a narrow-passage setting for the planning optimizer and consequently, leads it to a “timid” grasp in the execution phase \cite{grasp_timid_daring}. Soft materials on the other hand, gives the hand more freedom to have more contact points (even infinite number of points) than just a few points on the fingertips in the rigid hand case. Moreover, by exploiting the environment interactions \cite{rbo_thesis, PSH}, it helps reduce the complexity of the optimization problem and altogether results in a more comprehensive, more “darling” grasp \cite{grasp_timid_daring}. 

According to \cite{data_driven_survey}, depending on the priori knowledge, empirical methods can fall into either known objects, familiar objects or unknown objects category. While under “known objects” or “unknown objects” categories, basic structure of the objects or the hand were exploited to design grasp strategy. In familiar objects, the main idea is “similar objects will be treated similarly” and to determine “how similar”, a great number of authors proposed to incorporate machine learning techniques in their research. Classification algorithms such as SVM \cite{SVM_grasp} or ANN \cite{ANN_grasp} were applied to distinguish between graspable and non-graspable objects where inputs are 3d features extracted from point cloud information. Recently, after the emergence of deep learning, in the object recognition and segmentation fields, Convolutional Neural Network (CNN) is the most effective neural architecture choice. It dominated other non-deep learning approaches in almost any factors including performance, scalability and ease-to-use. Pinto et al. \cite{pinto_cnn} was the first and state-of-the-art work employing CNN for grasping and just two years after, C. Choi et al. \cite{cnn3d} applied similar ideas on a soft gripper. Apart from learning approaches for the perception phase, grasp synthesis, another branch of learning to grasp, aims to synthesize grasp from their own experience (trial-and-error) or human demonstration (actual human grasping). Kroemer et al. \cite{kroemer_kernel} proposed to use kernel-based nonparametric learning technique to construct a set of representations for primitive motions that can take object point clouds as input. This perception to primitive actions scheme has been widely applied in soft hand grasping as unlike rigid hand cases, thanks to the design, control strategy for soft hand is often simpler and mostly relies on primitive actions. Recently, Santina et al. \cite{santina_deep_grasp} and Gabellieri et al. \cite{gabellieri_deep_grasp} both have come up with a comprehensive controller for Pisa/IIT Soft Hand 2 robot that outputs direct motor controls from visual information. However, the former approach, as relies on a deep classifier, requires a large number of data whereas, the latter design is a non-deep learning approach, though more complicated with many hand-crafted components inside, needs a significantly less number of demonstrations.

From this short introduction, we start with the discussion of the hardware implementation of two popular soft hands: RBO and Pisa/IIT Soft hands in Section~\ref{sec:hardware}. Then in Section~\ref{sec:analytic}, we will briefly discuss the analytic approaches for soft hands controlling. Taking another perspective, Section~\ref{sec:data-driven} will explain different data-driven approaches including both the traditional and modern approaches based on deep learning. Finally, we conclude the letter in Section~\ref{sec:conclusion}.

\section{Mechanical design of soft hands}
\label{sec:hardware}

\subsection{RBO}
\label{sec:hardware_rbo}

\begin{figure}[H]
    \centering
    \includegraphics[width=0.5\textwidth]{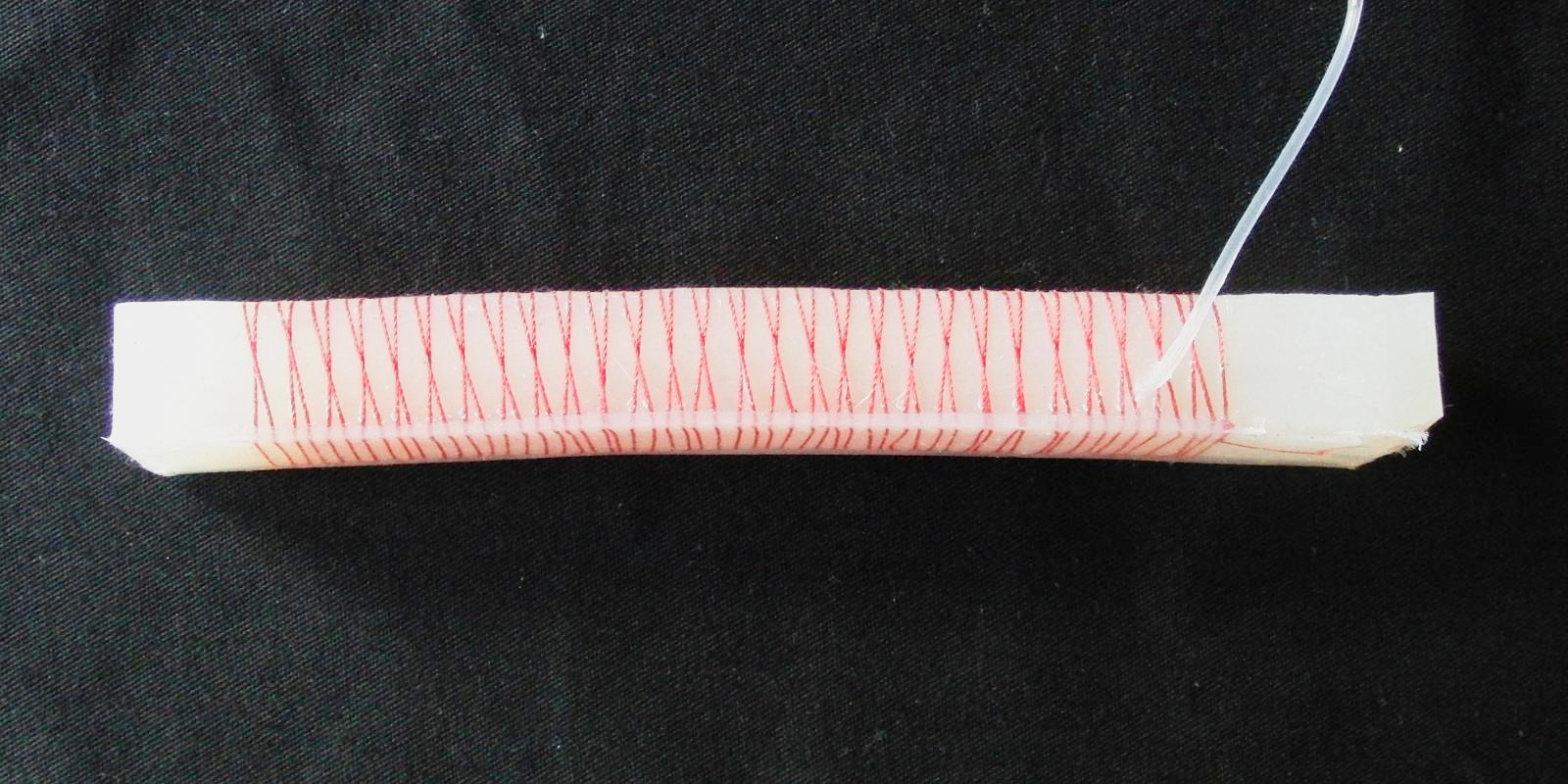}
    \caption{PneuFlex actuator \cite{rbo}}
    \label{fig:pneulex}
\end{figure}

Uncertainty in the real world poses a difficult problem for robotic hand designing. To deal with that, traditional rigid hands often require multiple highly accurate sensors. Meanwhile, through direct interaction with the targeted object and also the surrounding environment, humans can largely reduce the uncertainty to come up with a strong and reliable grasp. This environmental interaction capability along with the dexterity of human hands inspires many researchers to integrate soft material in their robot hand design. The RBO2 \cite{rbo_thesis} is one of the notable designs, it is an anthropomorphic hand with four fingers and one thumb, all made of silicone material. There are seven actuators in total, five corresponds to five fingers including the thumb and two extras inside the palm. These two actuators help the movement of the thumb more flexible and dexterous. The underneath mechanism they used is called the PneuFlex actuator, which is shown in Figure~\ref{fig:pneulex} below

The operation principle of the PneuFlex actuator is based on the deflation and inflation of a thin silicone tube installed inside \cite{rbo}. This air valve mechanism helps control the actuator ratio (curvature w.r.t applied pressure) and stiffness (change of curvature w.r.t change of moment). In addition, to avoid unexpected expanding, they helically twisted a thread wound around the soft finger and embedded a flexible mesh at the bottom side. This deformable  design gives the hand more freedom to exploit the environmental contacts. It is low inertia, robustness to different kinds of collisions and safe for the environment. Though many advantages have been discussed, RBO2 still has some limitations. First, the underlying mechanism based on the pneumatics is harder to accurately control the forces than the electromechanical systems normally employed in rigid hands. Secondly, at the current version, all the actuators of RBO2 have fixed stiffness and this limits the strength variability of a grasp.

\subsection{Pisa/IIT Soft hand}
\label{sec:hardware_psh}

In neuroscientific, researchers discovered that humans decide a grasp based on a set of patterns called postural synergies. This set helps simplify the robotic design significantly. In fact, in recent research, an underactuated hand capable of reproducing these synergies is more preferable than a fully actuated hand with many degrees of freedom. Pisa/IIT Soft hand is one of the successful hands following this idea. It has only one actuator but is able to control up to 19 joints to form the first few postures. It has been empirically proven that these few postures are sufficient to get a reliable grasp with a large variety of objects \cite{PSH}. Mathematically speaking, the connection between joints and the posture can be seen as follow
\begin{align*}
    q = S\sigma
\end{align*}
where $q$ is current joint position, $\sigma$ is the representative synergy and the synergies matrix $S \in R^{\#q \times \#\sigma}$. To convert this high level idea to a real design, they introduced a new concept called “adaptive synergies”. This specific scheme aims to control all 19 joints $q$ by using only one single actuator via the well-known differential transmission based on tendons and pulleys. Let $s \in R$ be a variable represented for the displacement of the extremities of the tendon, $\eta \in R$  is the motor torque of the only actuator, and $\tau \in R^{19}$ is joint torque. Then, the relation of these variables can be described in this equation 
\begin{align}
    s &= Rq \\ 
    \tau &= R^T \eta + K_qq
    \label{eq:psh_dyn}
\end{align}
where $R$ and $K_q$ are respectively called adaptive synergies and joint stiffness matrices \cite{grasp_timid_daring}. This $K_q$ matrix is employed to incorporate the elasticity effect of the elastic band into the system, an important part of the tendon-pulley mechanism. 

\begin{figure}[H]
    \centering
    \includegraphics[width=0.5\textwidth]{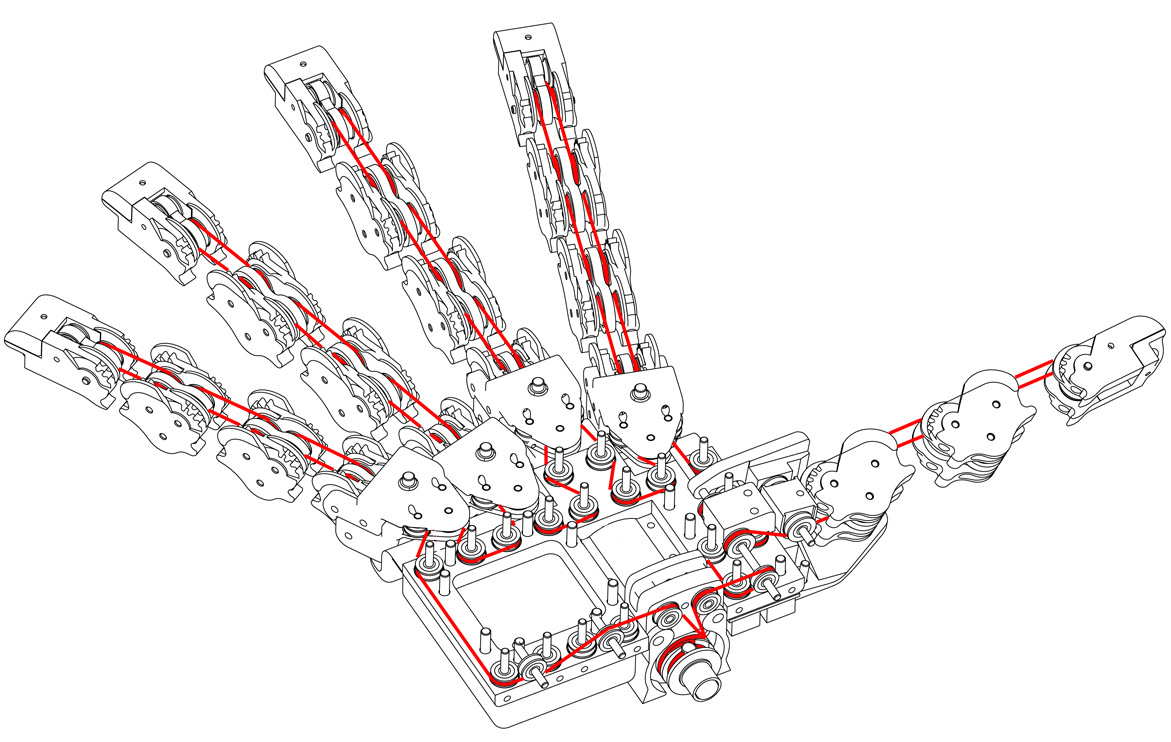}
    \caption{Pisa/IIT Softhand tendon routing \cite{PSH}}
    \label{fig:tendon_routing}
\end{figure}

As of RBO2, Pisa/IIT Soft hand design aims toward a strong and compliant grasp along with a capability of working safely under the cluttered environment. However, unlike RBO2, each finger of the Pisa/IIT Soft hand is rather “partially soft” with three to four phalanges joints in between. These joints are connected via an elastic ligament which is actually made from a polyurethane rubber. This turns out to be the main soft ingredient of the entire system. It helps the hand recover to default configuration after applying overexertion forces or being dislocated due to collision with a rigid obstacle. Throughout a specific tendon routing, all of  the joints are actuated by the passive anti-derailment pulleys shown in Figure~\ref{fig:tendon_routing}

\section{Analytic methods for soft grasping}
\label{sec:analytic}
Analytic approaches for autonomous grasping often exploit the hand dynamic and the geometric information of the targeted object to build a set of constraints for an optimization problem.  However, in practice, this information is only available in an approximated form. Lack of accurate sensors, disturbances from the model and uncertainties from the environment are the main reasons. Altogether, it significantly increases the difficulty of the original problem. In 2000, designing a grasp quality score that is robust to these unavoidable errors was an attractive topic \cite{analytic_grasping}. Roa and Suárez \cite{roa_suarez_icr} or Krug et al. \cite{Krug_icr} proposed efficient methods to deal with positioning errors. Their works mainly explored the concept of independent contact regions (iCRs) \cite{icr}, which is a set of regions such that as long as the robot hand makes contact with them, it will remain stable (i.e the grasp does not lose its force-closure property). In another perspective, instead of focusing on the object, they aim to explore the end-effector position robustness. To this end, caging information was heavily exploited. Rodriguez et al. Rodriguez et al. \cite{caging} showed that for a three-fingered gripper, there are some specific configurations (caging configurations) that can be used as waypoints for grasping a planar object. Once the manipulator stays in that configuration, the grasp remains stable without concerning the inaccuracy of the fingers position. 

\begin{figure}[H]
    \centering
    \includegraphics[width=0.7\textwidth]{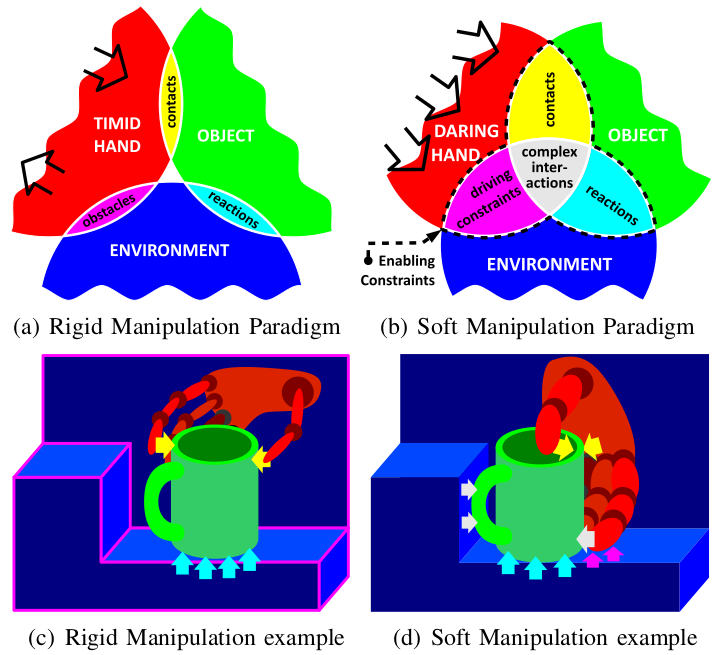}
    \caption{Comparison between rigid and soft hand \cite{PSH}}
    \label{fig:timid_daring}
\end{figure}

Although traditional methods give feasible solutions for a large set of grasping tasks, due to the fragility of the mechanical hand, the resulting grasp is often “timid” (i.e only the fingertips make actual contacts to the object). Figure~\ref{fig:timid_daring} (top-left) shows that the interaction capability of rigid hands is very limited. This passes the heavy workload to the planning scheme. More complicated constraints are needed to describe the environment, and hence force the planner to find the trajectory in a narrow passage. Meanwhile, in the top-right of Figure~\ref{fig:timid_daring}, the compliant structure of a soft manipulator helps relax the complexity of constraints, making it able to exploit the environmental contacts to form a more “daring” grasp. Similar to rigid hands planning, using simulation to design grasp strategies for soft hands is also a popular choice. Bonilla \cite{grasp_timid_daring} designed a simulation software for Pisa SoftHand based on the multi-body system framework ADAMS \cite{adams}. Thanks to the hardware implementation, the tendon driven kinematic of the hand can be mathematically described under a mass-spring system with multiple bodies. Specifically, as illustrated in Equation~\ref{eq:psh_dyn}, the final torque depends both on the current position of the joint (body) and its elastic band (spring) effect (represented by joint stiffness matrix $K$). One notable benefit of using ADAMS is that Coulomb friction at the joint level can be modelled by tuning the damping term of the rotation springs. This makes the simulation become more realistic (e.g remove some unphysical oscillations due to numerical errors) which would be useful for adapting the resulting grasping strategy to real physical SoftHand. Another simulation for SoftHand based on Klamp’t simulator also demonstrated a promising result \cite{psh_klamp}. Compared to the former ADAMS based simulation, it resolves the collision between hand and objects more effectively where less penerations are observed. This is the consequence of the combination between three advanced techniques: boundary-layer expanded meshes (BLEM), contact point clustering and adaptive time stepping. 
As opposed to Pisa SoftHand, RBO2 is a soft continuum hand where each finger is fully deformable. Simulating deformable objects is more complicated than rigid bodies-based systems. It requires to solve a set of Partial Differential Equations (PDE), and most of the numerical methods for solving PDE are computationally expensive. Finite Element Methods (FEM) is the most popular choice amongst others for deformable objects as it often results in a highly accurate and realistic solution, which is particularly suited for soft robotics. However, the cost of computation remains challenging. In a recent research, 	Pozzi et al. \cite{FEM_BRO} observed that the accuracy of FEM can be traded to increase the speed of simulation. Their idea is using the accurate but slow FEM to approximate a series of kinematic chains. In other words, transforming the problem of simulating soft continuum objects to simulating mass spring systems which can then be efficiently run under any fast and accurate multi-body based simulation. In this setting, they aim to “discretely” approximate the stiffness matrix from samples collected by running FEM in the preparation phase. As FEM is only needed in the first few hours of preparation, this approach significantly reduced the computation cost while still keeping the accuracy sufficiently high. 

Model-based methods give us a better understanding of the real world behaviors. However, the gap between the mathematical model and the real physical world remains large. In practice, due to the approximated manner of the numerical methods, highly non-linear movements pose a severe problem for simulation software. Also, uncertainty of nature is difficult to simulate and often ignored in the simulator. Hence, instead of encapsulating everything from nature into the simulator, in data-driven approaches, which will be discussed in the next section, they either take advantage of real world data to compensate for the disturbances of the model or even let the robot wildly move in the real environment to form a grasp from their own past experiences without the need of model.

\section{Data-driven methods for soft grasping}
\label{sec:data-driven}

Leveraging empirical data to enhance grasp quality is an attractive topic. In this section, we will discuss two popular directions, discriminative approaches and model-free grasp synthesis. 

\subsection{Discriminative approaches}

Observing that objects shared similar characteristics can be grouped together, it is reasonable to apply a single grasp strategy to the whole category of object. Determining the similarity between objects and generalizing it to similar but unseen objects is the ultimate goal of learning algorithms. In discriminative approaches for grasping, given visual information of objects, a “supervised learning” model is trained on a set of known objects with similar structure along with their labels (e.g graspable or contact points, same grasp synergies can be applied). Then in the testing phase, with only limited information of the new object (raw point cloud data), the same model can be applied. 

Before the marriage of deep learning and computer vision, low dimensional superquadric (SQ) representations for 3D objects were a popular choice for input of a supervised model. Given a 3D model, a traditional computer vision algorithm is applied to segment it into parts. Each part is then represented by a corresponding SQ. Depending on the purposes of the supervised model, SQ information is exploited differently. For example, if the goal is to determine which part of the object creates a stable grasp, force-closure points will serve as a label data on each SQ \cite{ANN_grasp}. Whereas in \cite{SVM_grasp}, an SVM algorithm is trained to maximize the grasp quality with SQ parameters and a grasp configuration as inputs. Although good results were observed, it remains unclear whether these methods are able to generalize into real scenarios where the information is available under approximated form (e.g point cloud data).

Deep learning based approaches, on the other hand, allow us to take high dimensional data as inputs. In 2015, Pinto et al. \cite{pinto_cnn} published the state-of-the-art results for grasping tasks based on Convolutional Neural Network (CNN). In this work, they aim to predict grasp locations directly from an RGB image. From 50k training datapoints being self collected by a Baxter robot over 700 hours, the robot can detect the correct graspable points with accuracy up to 79.5\% on the test dataset with sufficiently high diversity of newly unseen objects (none of the traditional approaches can achieve this back then). In 2018, C. Choi et al. \cite{cnn3d} came up with a similar idea for soft grasping. They presented a CNN network that takes raw 3D point cloud data of an object and outputs a suitable caging grasp for it. As mentioned earlier in the previous sections, to compose a stable grasp for a soft hand, we do not need to detect the exact contact points, but rather focus on finding a right posture for the grasping object. Under this 3D CNN model, the percentage of successful grasping rises up to 87\% on a test set with multiple unknown objects. Though both of the results look promising, the presented deep networks were only tested on a simple - either rigid or soft - gripper. Visar et al. \cite{failure_grasp_detection} moved one step beyond that, they proposed to apply deep learning on anthropomorphic soft hands, the Pisa/IIT SoftHand and RBO. Two different deep networks were introduced in their work. Given a set of completed grasps, the first one is trained to discriminate failed grasps from the successful ones. This data is used as ground truth labels for training the second network which is also their main contribution. It aims to predict failure events in a few seconds in the future before it actually happens. This pre-event detection mechanism would leave the planning scheme enough time to react and give a better plan. Unlike \cite{pinto_cnn} and \cite{cnn3d}, the input for both of the two networks are accelerations and velocities sensing from IMU information distributed on the soft hand. This kind of information is well suited for soft robotics since measuring forces is extremely difficult for soft material due to its deformable properties. In the paper, they showed that the second network can detect 100\% grasp failure in average 1.96 seconds before the actual events.

\subsection{Model-free grasp synthesis}

Grasp synthesis is a data-driven method aimed to build a database of pre-grasps from empirical data. For future grasps, it can decide which grasping strategy to apply by querying from the database. Following this idea, \cite{allen_group_1} and the follow up works \cite{allen_group_2, semantic_grasping} build a grasp database based on the concept of Eigengrasp planner. Observing that the object features (pose and shape) and the grasp posture has a strong connection, i.e same synthesized grasp could be applied if the targeted object with similar shape are placed at a similar pose. H. Dang et al. \cite{semantic_grasping} showed that it is very efficient to find an optimal grasping path for a novel object (with known 3D mesh and belongs to one of the defined categories) under a low dimensional space of grasp posture (Eigengrasp space). Projecting high dimensional inputs into low dimensional subspace helps reduce the number of synthesizing grasps, only a few trials of the same category object are needed to construct the database. Also, in the inference phase, even if the object has not been inserted into the database yet, via interpolation on the subspace it is highly possible to find a suitable grasp posture for that. 

Due to the uncertainty of nature, knowing exactly the pose and shape of the object is not always possible. In one of the mentioned approaches, C.  Goldfeder et al. \cite{allen_group_2} designed a framework to deal with partial observation data. It takes direct information from real sensors, processes it into a queryable shape and matches it with the most similar 3D model in the database. However, in this setting, the grasp is still being constructed in an offline fashion, i.e grasp posture is only available for a known model. There is another direction for grasp synthesis without the need to build a database of models. They instead aim to learn a function that maps directly raw visual input into primitive actions. O. Kroemer et al. \cite{kroemer_kernel} designed such a learning scheme based on trial and error. Instead of considering the whole object, they only take into account sub parts of the object that a grasp can be constructed on. The direct perception action mapping is formed by a set of kernel bases. Its aim is to detect how similar the subparts are and from that output a suitable action. Because the entire process is trial-and-error, only a single demonstration from a human for each primitive action is needed. 

Regarding the grasp synthesis method for soft hands, Santina et al. \cite{santina_deep_grasp} designed a comprehensive framework to achieve grasp from raw sensors. Taking the inspiration of human grasping, a grasping task can be constructed from three levels of abstractions. In the highest level, a deep classifier is trained to return a probabilistic distribution over a set of motion primitives. Taking an RGB image from a camera, a well known network architecture for object detection YOLO is employed to get the location of the targeted object. This information is then propagated to the pretrained Inception network with a modification on the last fully-connected layer for outputting softmax distribution of primitive actions. In the middle level, a closed loop control inspired from humans is employed to react appropriately given sensor information from the IMUs. Acting is the lowest level of this framework and it completely counts on the embodied intelligence of the hand. The compliant structure of the Pisa/IIT Softhand itself helps reduce the local uncertainties and hence mitigate the gap between the planned and the actual grasp. The proposed framework was extensively tested on real hardware under 20 objects, within 111 conducted grasps, 81.1\% was successful. Although the final outcome of this work is impressive, as mostly relying on deep neural networks, it is hard to explain what the problem is if a grasp fails. Gabellieri et al.  \cite{gabellieri_deep_grasp} follows the same hierarchical structure and proposed another framework that combines multiple handcrafted algorithms. As opposed to deep learning, the demand for data of handcrafted algorithms is significantly lower and more importantly, it is possible to know the reason behind a failure grasp. In this work, point cloud data of the object obtained by an RGB-D camera is fed into the MVBB algorithm to get a set of cuboid boxes. Cuboid boxes are the principle representation for an unknown object which is often employed in rigid body collision detection and handling. These boxes are then sorted and the candidate with the highest graspable potential is then selected as a feature for the Decision Tree Regression (DTR) to decide which poses of it to make a grasp. This information is then passed  to the grasp selection block to select a suitable grasp posture. Inside the grasp selection block, it takes the grasp pose, check for possible collisions, select an initial grasp, check whether the grasp is kinematically feasible and if it is not, choose another grasp otherwise pass them to the robot and execute. Similar to \cite{santina_deep_grasp}, they tested the proposed framework on real grasping tasks, achieving a successful rate 86.7\% on 105 grasps of 21 objects. Interestingly, the whole framework is a combination of “whitebox” components, hence it is possible to investigate what actually happens during the grasp.

Apart from finding a suitable primitive action for soft hands, A. Gupta et al. \cite{rbo_gps} proposed a data-driven method that can directly output raw commands for each joint. They use the RBO soft hand to test their method, which has seven degrees of freedom. As discussed in Section~\ref{sec:hardware_rbo}, the dynamic for this soft robot is not well-defined, hence in this work, they aimed to learn both the dynamic and the policy in an iterative fashion through a learning framework called Guided Policy Search (GPS) \cite{gps}. GPS idea is learning a global policy from multiple local policies. Each local policy is a simple Linear Quadratic Regulator (LQR) control. Since LQR requires model knowledge which is not available for the RBO soft hand, they instead use Gaussian Mixture Model (GMM) to approximate each local dynamic and learn it from samples. The whole framework is thus completely model-free, the user just needs to define an appropriate task-related objective function, then the robot can learn to complete the task on its own. Interestingly, GPS needs less samples to converge to an optimal policy compared to other model-free methods as due to the usage of the GMM dynamic and the LQR, the search space for finding the final global policy is much smaller. In the report, they showed that the RBO can learn different manipulation skill sets beyond grasping: valve turning and moving beads on an abacus.

\section{Conclusion}
\label{sec:conclusion}
We have discussed a broad overview for autonomous soft hand grasping in this short letter. Unlike rigid hands, soft hands often result in a strong and reliable grasp. Thanks to the compliant structure in the design (e.g the tendon in Pisa/IIT soft hand or the silicon material in RBO), it helps reduce local uncertainty and enables the hand to exploit the environment contacts rather than avoiding it. Although the hardware implementation of soft hands brings a lot of benefits, controlling a soft hand in an autonomous fashion is challenging. This is the result of sensing capabilities limitation and highly non-linear dynamic. As opposed to rigid hands, analytic methods for soft hands do not aim to find a set of contact points but rather simulate the whole grasping process in a simulator. Depending on the type of the soft materials, the simulation system is either a multi-bodies or a fluid-based system. Nonetheless, none of the simulation softwares can capture all the real world aspects.  Data-driven is another approach that exploits empirical data to either encounter with the real world uncertainty or directly learn a suitable strategy for grasping. The latter does not even need a model, their objective is to learn a model that outputs a grasping plan directly from raw sensor information. Although different solutions have been proposed in a few recent years, autonomous grasping remains a challenge. Only a limited number of objects have been tested and the number of works involved grasping in a cluttered environment is not high either, which is still far behind human grasping.


%
\bibliography{my}

\end{document}